\pdfoutput=1
\documentclass[11pt]{article}

\usepackage[]{acl}

\usepackage{times}
\usepackage{latexsym}

\usepackage[T1]{fontenc}
\usepackage[utf8]{inputenc}

\usepackage{graphicx}  % [Chen WU] Added and checked.
\usepackage{mathtools}  % [Chen WU] Added and checked.
\usepackage{amssymb}  % [Chen WU] Added and checked.
\usepackage{amsthm}  % [Chen WU] Added and checked.
\usepackage{multirow}  % [Chen WU] Added and checked.
\usepackage{subfigure}  % [Chen WU] Added and checked.
\usepackage{booktabs}  % [Chen WU] Added and checked.
\usepackage{algorithm}  % [Chen WU] Added and checked.
\usepackage{algcompatible}  % [Chen WU] Added and checked.
\usepackage{bm}  % [Chen WU] Added and checked.
\usepackage{xcolor}  % [Chen WU]  Added and should not be used in the final version.
\usepackage{colortbl}  % [Chen WU] Added and checked.
\newcommand{\itm}[1]{\textrm{\textit{#1}}}  % [Chen WU] Added and checked
\usepackage{eqparbox}  % [Chen WU] Added and checked.
\usepackage{xspace}

\usepackage{microtype}

\iffalse
    \newcommand{\an}[1]{\textcolor{cyan}{}}
    \newcommand{\rui}[1]{\textcolor{blue}{}}
    \newcommand{\ziming}[1]{\textcolor{red}{}}
    \newcommand{\chen}[1]{\textcolor{purple}{}}
    \newcommand{\budha}[1]{\textcolor{green}{}}
    \newcommand{\ys}[1]{\textcolor{brown}{}}
    \newcommand{\dr}[1]{\textcolor{magenta}{}}
    \newcommand{\revise}[2]{\textcolor{red}{For #1: #2}}
\else
    \newcommand{\an}[1]{\textcolor{cyan}{\bf\small [AN: #1]}}
    \newcommand{\rui}[1]{\textcolor{blue}{\bf\small [Rui: #1]}}
    \newcommand{\ziming}[1]{\textcolor{red}{\bf\small [Ziming: #1]}}
    \newcommand{\chen}[1]{\textcolor{purple}{\bf\small [Chen: #1]}}
    
    \newcommand{\budha}[1]{\textcolor{green}{\bf\small [Budha: #1]}}
    \newcommand{\ys}[1]{{\color{brown}{\small{\bf [Yusen: #1]}}}}
    \newcommand{\dr}[1]{{\color{magenta}{\small{\bf [Drago: #1]}}}}
    \newcommand{\revise}[2]{\textcolor{black}{#2}}
\fi

\newcommand{\method}{\textsc{Dyle}\xspace}
\usepackage{pifont}
\newcommand{\cmark}{\ding{51}}
\newcommand{\xmark}{\ding{55}}
\newcommand{\eg}{\textit{e.g., \xspace}}

\title{
\textsc{Dyle}: Dynamic Latent Extraction for Abstractive \\ Long-Input Summarization
}

\author{
Ziming Mao\thanks{\ \ Equal Contributions.} $\ ^{1}$
\quad Chen Henry Wu\footnotemark[1] $\ ^{2}$
\quad Ansong Ni$^{1}$
\quad Yusen Zhang$^{3}$
\\{\bf 
\quad Rui Zhang$^{3}$
\quad Tao Yu$^{4}$
\quad Budhaditya Deb$^{5}$}
\\{\bf 
\quad Chenguang Zhu$^{5}$
\quad Ahmed H. Awadallah$^{5}$
\quad Dragomir Radev$^{1}$} 
\\
$^1$ Yale University 
\quad $^2$ Carnegie Mellon University
\quad $^3$ Penn State University 
\\
\quad $^4$ The University of Hong Kong
\quad $^5$ Microsoft Research
\\
\tt{ziming.mao@yale.edu},
\tt{henrychenwu@cmu.edu}
}

\begin{document}
\maketitle

\begin{abstract}
Transformer-based models have achieved state-of-the-art performance on short-input summarization. However, they still struggle with summarizing longer text. 
In this paper, we present \method, a novel dynamic latent extraction approach for abstractive long-input summarization. 
\method jointly trains an extractor and a generator and treats the extracted text snippets as the latent variable, allowing dynamic snippet-level attention weights during decoding. To provide adequate supervision, we propose simple yet effective heuristics for oracle extraction as well as a consistency loss term, which encourages the extractor to approximate the averaged dynamic weights predicted by the generator. 
We evaluate our method on different long-document and long-dialogue summarization tasks: GovReport, QMSum, and arXiv. Experiment results show that \method outperforms all existing methods on GovReport and QMSum, with gains up to 6.1 ROUGE, while yielding strong results on arXiv.
Further analysis shows that the proposed dynamic weights provide interpretability of our generation process.\footnote{Our code is available at: \url{https://github.com/Yale-LILY/DYLE}
}
\end{abstract}

\section{Introduction}
Transformer-based \cite{VaswaniSPUJGKP17} pretrained language models (PLMs) such as BART \cite{LewisLGGMLSZ20} and T5 \cite{RaffelSRLNMZLL20}, have achieved state-of-the-art performance on short text summarization. However, due to the high memory complexity of the full self-attention \cite{tay2020long}, PLMs still struggle to handle long inputs \cite{rohde2021hierarchical}. \textit{Model efficiency} and \textit{summary quality} present a pair of challenges \cite{HuangCPJW21}: models need to capture information scattered across the long input while maintaining a low computational cost.

%%%%%%%%%%%%%%%%%%%%%%%%%%%%%%%%%%%%%%%%%%%%%%%%%%%%%%%%%%%%%%%%%%%%%%%%%%%%%%
\begin{figure}[t]
	\centering
	\includegraphics[width=0.95\linewidth]{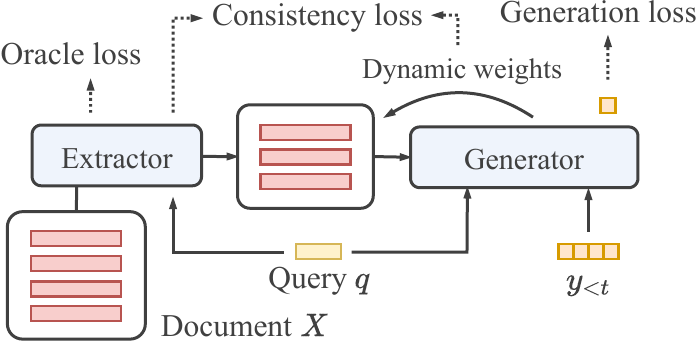}
	\caption{\label{fig:overview} Overview of our approach. The input is a document $X$ (each $x \in X$ is a sentence) and an optional query $q$, and the output is a summary $y$. 
	}
\end{figure}
%%%%%%%%%%%%%%%%%%%%%%%%%%%%%%%%%%%%%%%%%%%%%%%%%%%%%%%%%%%%%%%%%%%%%%%%%%%%%%

Prior models tackled long input summarization mostly in four ways. First, \textit{sparse attention} \cite{Child19Sparse,Beltagy20Longformer,Tay20Survey} is used to reduce the memory complexity of the Transformers so that they can attend to more tokens. 
Second, \textit{extract-then-generate} methods extract salient texts from the input and then summarize the extracted texts. Extractors are either independently trained with full supervision \cite{ZhongYYZMJACLQR21} 
or optimized using reinforcement learning \cite{Williams92, BansalC18, bae2019summary, bravzinskas2021learning}. 
Third, models are proposed to divide source text into sections \cite{Gidiotis2020Dancer, wu2021recursively,liu2021end} which are individually summarized and combined to form a full summary.
Fourth, hierarchical models \cite{rohde2021hierarchical, zhu2020hierarchical} 
improve summarization by capturing sentence or discourse level dependencies. We elaborate on these four directions and their limitations in Section \ref{related-work}.

We believe that the extract-then-generate approach mimics how a person would handle long-input summarization: first identify important pieces of information in the text and then summarize them \revise{R1}{\cite{kiyoumarsi2015evaluation,sun2020summarize}}. The extract-then-generate framework is based on the assumption that salient information useful for summarization only occupies a small portion of the input, which is a sensible assumption given the long input length. This approach shortens the source input to a pre-set length, which addresses the main challenge of the model not being able to handle longer input beyond a certain limit. However, previous separately-trained extract-then-generate approaches are limited as they 
suffer from cascaded errors from the extractor to the generator. Though various reinforcement learning techniques are introduced to bridge the two steps, they have noticeable drawbacks (discussed in Section \ref{subsec:dynamic-latent-extraction}), and we argue that the long input makes this approach suboptimal.

In this paper, we propose a new approach for long-input summarization: \textit{\underline{Dy}namic \underline{L}atent \underline{E}xtraction for Abstractive Summarization} (\textsc{Dyle}).
\method \textit{jointly trains} the extractor and the generator and keeps the extracted text snippets latent.
For an output token, \method compute its probability conditioned on each input snippet \textit{separately}, and its generation probability is computed by \textit{marginalizing} over all the input snippets under a learned \textit{dynamic weights} assigned by the generator conditioned on the previously generated tokens. 

We optimize the extractor with two surrogate losses. First,  we compute the \textit{extractive oracle} based on the reference summary with a greedy search over the best ROUGE scores. These oracle snippets are used as targets for the extractor learning signal.
Moreover, we propose \textit{consistency loss} to encourage the extractor to approximate its own predicted weights on the snippet to the averaged dynamic weights predicted by the generator. 

We conducted experiments on three long-input summarization datasets: GovReport \cite{HuangCPJW21} and arXiv \cite{cohan-etal-2018-discourse} for long-document summarization,  and QMSum \cite{ZhongYYZMJACLQR21} for long-dialogue summarization. 
Our method \method largely outperforms existing methods on GovReport and QMSum, while achieving strong results on arXiv. Notably, \method yields gains of 4.2/6.1/4.0 of ROUGE-1/2/L points over the previous best method on GovReport. 
These experiments demonstrate the generalizability of \method to multiple long-input summarization tasks.

We summarize our contributions as follows:
\begin{itemize}

    \item We introduce \method, a dynamic latent extraction approach for abstractive long-input summarization. \method better captures information in the long input and reduces computational cost;
    \item We propose multiple auxiliary optimizations for the effective training of \method: 1) extractive oracle as a learning signal for the extractor; 2) consistency loss that bridges extraction and generation; 3) hybrid training methods that make the extraction more robust;
    \item Experimental results show that \method largely outperforms the state-of-the-art on two long input summarization datasets. We also conducted a detailed analysis that shows dynamic weights improve model interpretability.
\end{itemize}

\section{Related Work} 
\label{related-work}

\revise{R3}{We introduce in detail the four main categories of methods in recent work to address long-input summarization tasks.}

\paragraph{Sparse attention mechanism} The full attention mechanism has a quadratic memory cost. Prior research works have proposed different sparse attention mechanisms to reduce the memory cost. Longformer \cite{Beltagy20Longformer} uses a dilated sliding window of blocks and global attention patterns. BigBird \cite{zaheer2020big} employs sliding windows and random blocks. Reformer \cite{Kitaev2020Reformer} uses the locality-sensitive hashing. In addition to optimizing the encoder self-attention, \citet{HuangCPJW21} proposes head-wise positional strides to reduce the cost of the encoder-decoder attention. However, sparse attention diminishes the benefits of pretraining and sacrifices parts of the receptive field.

\paragraph{Extract-then-generate method} This method extracts salient text snippets from the input, followed by generating an overall summary. Most of these approaches are trained separately \cite{zhang2019pretraining, lebanoff-etal-2019-scoring, xu-durrett-2019-neural, BajajDKKUWBDDM21, zhang-etal-2021-exploratory-study}, which suffer from information loss as we pass the extracted snippets to the generator. Some approaches attempt to reduce that loss by bridging the two stages. \citet{BansalC18} adopts reinforcement learning (RL) with a sentence-level policy gradient. \citet{bae2019summary} proposes summary-level policy gradient. Using RL suffers from various drawbacks on long input texts, which will be elaborated in Section \ref{subsec:dynamic-latent-extraction}. \textsc{Dyle} is different as we jointly train an extract-then-generate model for summarization using latent variables. 

\paragraph{Divide-and-conquer approach} A common approach in long input summarization is divide-and-conquer \cite{Gidiotis2020Dancer, grail-etal-2021-globalizing, summn}. It breaks a long input into multiple parts, which are summarized separately and combined to produce a final summary. However, these models do not capture the contextual dependencies across parts and assume that the input has certain structure.

\paragraph{Hierarchical models} Various hierarchical models have been proposed to handle the longer inputs. \citet{cohan-etal-2018-discourse} models the document discourse structure with a hierarchical encoder and a discourse-aware decoder. HAT-Bart \cite{rohde2021hierarchical} proposes a new Hierarchical Attention Transformer-based architecture that attempts to capture sentence and paragraph-level information. HMNet \cite{zhu2020hierarchical} builds a hierarchical structure that includes discourse-level information and speaker roles. However, these models focus mainly on model performance and not on reducing the memory and computational cost.

\section{Our Approach}

An overview of our approach is shown in Figure~\ref{fig:overview}. In Section~\ref{subsec:formulation}, we formulate our task and the extractor-generator framework. In Section~\ref{subsec:extractor}, we introduce our parameterization of the extractor for long inputs.
In Section~\ref{subsec:dynamic-latent-extraction}, we introduce generator formulation and the novel consistency loss. 
The extractor module is both optimized with the consistency loss and the oracle loss, which we elaborate on in Section~\ref{subsec:oracle-loss}. The overall training objective is summarized in Section~\ref{subsec:overall-objective}. 

\subsection{Extractor-Generator Framework}
\label{subsec:formulation}
In the long-input summarization task, the input consists of $L$ text snippets, $X = (x_1, \ldots, x_L)$, and an optional query $q$ if a query is paired with a summary. In long-input summarization, the number of text snippets, $L$, could be potentially large.
The output is a summary $y$ of length $T$. For the dialogue summarization task, dialogue utterances by each speaker are used as snippets. For documents, we tokenize the input into sentences and use each sentence as a snippet. 
The goal is to learn a model that generates a sequence of summary tokens $y$ given the input snippets $X$ and the previously generated tokens $y_{<t}$: 
\begin{displaymath}
    P_{\theta}(y|q, X) = \prod_{t=1}^{T} P_{\theta}(y_{t}|q, X, y_{<t})
\end{displaymath}

The \textit{extractor} 
takes the query and the source text as input and outputs a score $s_{i} = E_{\eta}(q, x_{i})$ 
for each text snippet $x_{i}$. Here $\eta$ is the extractor parameters. We extract $K$ snippets $X_{K}$ from the document $X$ based on their scores:
\begin{equation}
\label{eq:topk}
    X_{K} = \text{top-}K (E_{\eta}(q, x_{i}), x_{i} \in X)
\end{equation}
After retrieving $X_{K}$ from $X$, the extractor-generator framework models the output probability by replacing $X$ with $X_{K}$, i.e., 
\begin{equation}
\label{eq:dynamic-rag-topk}
\begin{split}
    P_{\theta}(y|q, X) &= P_{\theta}(y|q, X_{K}) \\
    &= \prod_{t=1}^{T} P_{\theta}(y_{t}|q, X_{K}, y_{<t})
\end{split}
\end{equation}
Note that the top-$K$ operation in Eq.~(\ref{eq:topk}) is non-differentiable, and we do not propagate gradients through top-$K$; instead, we propose methods to optimize the extractor in Section~\ref{subsec:dynamic-latent-extraction} and Section~\ref{subsec:oracle-loss}. 

%%%%%%%%%%%%%%%%%%%%%%%%%%%%%%%%%%%%%%%%%%%%%%%%%%%%%%%%%%%%%%%%%%%%%%%%%%%%%%
\begin{figure}[t]
	\centering
	\includegraphics[width=0.95\linewidth]{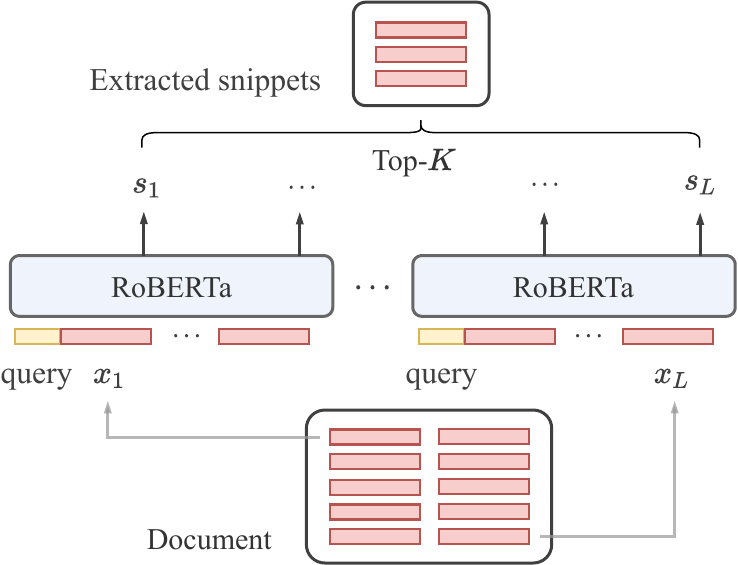}
	\caption{\label{fig:extractor} Long-input extractor. We divide the document into chunks, each containing consecutive snippets. A shared RoBERTa encodes each chunk independently. 
	}
\end{figure}
%%%%%%%%%%%%%%%%%%%%%%%%%%%%%%%%%%%%%%%%%%%%%%%%%%%%%%%%%%%%%%%%%%%%%%%%%%%%%%

\subsection{Extractor for Long Inputs}
\label{subsec:extractor}
An interesting research question is how to design the extractor for long inputs. Limited by GPU memory, it is impractical to concatenate all snippets and encode them with a large pre-trained language model. As shown in Figure~\ref{fig:extractor}, we group consecutive snippets into \textit{chunks}. We concatenate the query $q$ with each chunk and compute the encoded vector for each snippet independently within the chunk it belongs to. We project the encoded vectors to scalar scores $s_{i} = E_{\eta}(q, x_{i})$ using an MLP. 

\subsection{Generator with Dynamic Weights}
\label{subsec:dynamic-latent-extraction}
\paragraph{Challenges} 
An extract-then-generate model faces two challenges in long-input summarization. The first challenge is that the extraction operation (top-$K$ in Eq.~(\ref{eq:topk})) is non-differentiable.
One approach is to adopt RL-based optimizations \cite{BansalC18, bae2019summary}, which has two drawbacks. First, reinforcement learning for large action spaces (i.e., extracting $K$ out of $L$ snippets when $L$ is very large) has high variances. Second, current methods mostly use sentence-level ROUGE \cite{BansalC18} or summary-level ROUGE \cite{bae2019summary} as training rewards. Using sentence-level ROUGE could potentially select sentences with overlapping contents \cite{narayan2018ranking}, resulting in redundant final summaries. Using a summary-level ROUGE leads to the sparsity of the training signal, and longer input makes this approach harder to train. The second challenge is interpretability: one might want to know whether the generator is leveraging the extracted information at each decoding time step. 

To address these challenges, we propose a generator that dynamically assigns weights to every extracted snippet at each time step. Different from the extractor scores, which are independent of the decoding time step, the generator assigns different dynamic scores at different time steps. Dynamic weights make the decoding process interpretable and help denoise the extraction by down-weighting irrelevant snippets. It also provides training signals for the extractor using \textit{consistency loss}. 

%%%%%%%%%%%%%%%%%%%%%%%%%%%%%%%%%%%%%%%%%%%%%%%%%%%%%%%%%%%%%%%%%%%%%%%%%%%%%%
\begin{figure}[t]
	\centering
	\includegraphics[width=0.95\linewidth]{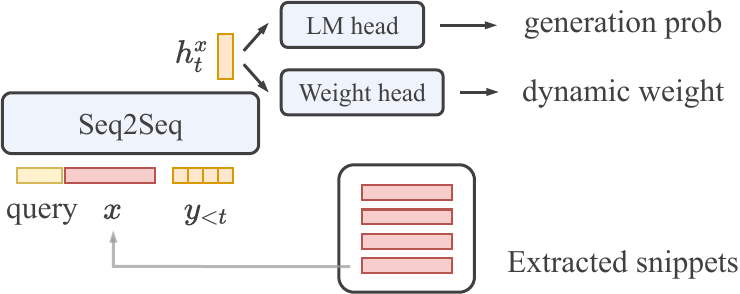}
	\caption{\label{fig:generator} At each decoding time step, our generator predicts the dynamic weight and the generation probability for each extracted snippet. }
\end{figure}
%%%%%%%%%%%%%%%%%%%%%%%%%%%%%%%%%%%%%%%%%%%%%%%%%%%%%%%%%%%%%%%%%%%%%%%%%%%%%%

\paragraph{Generator formulation} The overview of the generator is shown in Figure~\ref{fig:generator}. 
For each extracted snippet $x$, the generator predicts the generation probability $P_{\theta}(y_{t}|q, x, y_{<t})$ on this snippet and a dynamic weight $P_{\theta}(x|q, X_{K}, y_{<t})$ for this snippet. The independent encoding of each extracted snippet saves memory because the snippets do not need to attend to each other. Without loss of generality, we assume that $P_{\theta}(\cdot|q, x, y_{<t})$ is computed by first mapping the input $(q, x, y_{<t})$ to a contextualized representation vector $\bm{h}^{x}_{t}$. For Transformers \cite{VaswaniSPUJGKP17} and encoder-decoder with attention models \cite{BahdanauCB14}, $\bm{h}^{x}_{t}$ is usually the model's output before the final language model head. The generation probability $P_{\theta}(y_{t}|q, x, y_{<t})$ is computed by feeding $\bm{h}^{x}_{t}$ into the language model head. For the dynamic weight $P_{\theta}(x|q, X_{K}, y_{<t})$, we adopt a separate MLP to map each $\bm{h}^{x}_{t}$ to a scalar logit $l^{x}$, and $P_{\theta}(\cdot|q, X, y_{<t})$ is defined as $\mathrm{softmax}(\{l^{x}\}_{x \in X})$. We compute the generation probability by marginalizing over all extracted snippets:
\begin{equation}
\begin{split}
    &P_{\theta}(y|q, X_{K}) = \prod_{t=1}^{T} \sum_{x \in X_{K}} \\
    &\quad\quad\ \ \ P_{\theta}(y_{t}|q, x, y_{<t}) P_{\theta}(x|q, X_{K}, y_{<t})
\end{split}
\end{equation}
The dynamic weight $P_{\theta}(x|q, X_{K}, y_{<t})$ at each decoding time step $t$ allows us to interpret how the generator utilizes the extracted snippets. For example, a larger weight to a particular snippet indicates the larger importance of the snippet to the current decoding time step. 
The generation loss is defined as the NLL of the gold summary: 
\begin{equation}
\label{eq:s2s-loss}
    \mathcal{L}_{\itm{gen}}^{\theta} = -\log P_{\theta}(y|q, X_{K})
\end{equation}
where $P_{\theta}(y|q, X_{K})$ is defined in Eq.~(\ref{eq:dynamic-rag-topk}). Here we do not propagate gradients of $\mathcal{L}_{\itm{gen}}^{\theta}$ to the extractor parameters since top-$K$ is non-differentiable. Instead, methods to optimize the extractor are described in  Section~\ref{subsec:dynamic-latent-extraction} and Section~\ref{subsec:oracle-loss}.  

\paragraph{Consistency loss} We also leverage the dynamic weights to provide a training signal for the extractor. Since the dynamic weight of a snippet can be interpreted as the importance of the snippet at a particular time step, we average the dynamic weights over all the decoding steps and view the averaged weight as the overall importance of the snippet.  
Based on this intuition, we propose what we term as \textit{consistency loss}, which measures the distance between the averaged dynamic weights distribution and the extractor distribution. We want these two distributions to be close on an \textit{arbitrary} subset of $X$. For simplicity, we take $X_{K}$ as the subset and define the consistency loss as
\begin{equation}
\label{eq:consistency-loss}
\begin{split}
    &\mathcal{L}_{\itm{consist}}^{\eta} = \mathrm{KL}\Big[\frac{1}{T} \sum_{t=1}^{T} P_{\theta}(\cdot|q, X_{K}, y_{<t}) \ || \\
    &\quad\quad\quad\quad\quad \text{softmax}(E_{\eta}(q, x_{i}), x_{i} \in X_K)\Big]
\end{split}
\end{equation}
Note that the consistency loss is superscripted with the extractor's parameters $\eta$, which means that we do not compute gradients for the generator's parameters $\theta$. Since we want the distributional distance to be small on an \textit{arbitrary} subset of $X$, we do not propagate gradients through the top-$K$ operator. 

\subsection{Leveraging Extractive Oracles}
For long-input summarization, the extracted snippets $X_{K}$ used during training are important for stable optimization. Instead of using $X_{K}$ defined in Eq.~(\ref{eq:topk}), we propose to leverage \textit{extractive oracles} during training. No extractive oracles are used during test time.

\paragraph{Greedy search for extractive oracles} 
\textit{Extractive oracles} denote a set of selected text snippets whose concatenation maximizes the evaluation metric given the gold summary. We implement the extractive oracle using greedy search. Specifically, we start with an empty set, and we iteratively select a snippet from the input such that the concatenation of that snippet and the already selected snippets maximizes the average of ROUGE-1, ROUGE-2, and ROUGE-L scores given the gold summary. We denote the extractive oracles as $X_{o}$. 

\paragraph{Hybrid training} We leverage the extractive oracles to define $X_{K}$ used during training. If the number of oracles equals or exceeds $K$, we define $X_{K}$ as the first $K$ oracle snippets. If the number of oracles is less than $K$, we define $X_{K}$ as the union of $X_{o}$ and the top snippets ranked by the extractor that is not appearing in $X_{o}$. Such hybrid training has two benefits. First, compared with $X_{K}$ defined in Eq.~(\ref{eq:topk}), it provides higher-quality inputs to the generator. Second, it reduces the reliance on the oracle and improves the generalizability of our model beyond the training set, as other text snippets omitted in the greedy search might help the generation. 

\paragraph{Oracle loss} 
\label{subsec:oracle-loss}
The extractive oracles $X_{o}$ are used as a supervision signal for the extraction part of our model. The oracle loss $\mathcal{L}_{\itm{oracle}}^{\eta}$ is computed from the cross-entropy loss between all chunks in the extractor selected set and the extractive oracle.  Formally, the oracle loss is computed as
\begin{equation}
\begin{split}
    \mathcal{L}_{\itm{oracle}}^{\eta} = - \frac{1}{|X_{o}|}\sum_{x \in X_{o}} \log \frac{e^{E_{\eta}(q, x)}}{\sum_{x_{i} \in X} e^{E_{\eta}(q, x_{i})}}
\end{split}
\end{equation}

\subsection{Training Objective}
\label{subsec:overall-objective}
The overall training objective of our method is
\begin{equation}
\label{eq:overall-objective}
    \mathcal{L}^{\theta, \eta} = \lambda_{\itm{g}}\mathcal{L}_{\itm{gen}}^{\theta} + \lambda_{\itm{o}}\mathcal{L}_{\itm{oracle}}^{\eta} + \lambda_{\itm{c}}\mathcal{L}_{\itm{consist}}^{\eta}
\end{equation}
where $\lambda_{\itm{g}}$, $\lambda_{\itm{o}}$, and $\lambda_{\itm{c}}$ are hyperparameters to balance the loss components. Gradients are computed for the superscripted parameters. Specifically, the extractor is solely optimized with the consistency loss and the oracle loss, and the generator is solely optimized with the generation loss. 

\section{Experiment Setups}

\subsection{Datasets}
We consider the following long-input abstractive summarization datasets as evaluation benchmarks:\footnote{QMSum and arXiv can be accessed through SummerTime \cite{ni-etal-2021-summertime}.}
\paragraph{QMSum} \cite{ZhongYYZMJACLQR21} is a benchmark for query-based multi-domain meeting summarization. It consists of meetings from three domains: AMI \cite{carletta2005ami}, ICSI \cite{janin2003icsi}, and committee meetings of the Welsh Parliament and Parliament of Canada; % The meetings in this dataset comprise of a large number of turns uttered by multiple speakers. 

\paragraph{GovReport} \cite{HuangCPJW21} is a large-scale long document summarization dataset, consisting of about 19.5k U.S. government reports with expert-written abstractive summaries; GovReport is a good benchmark as it contains significantly longer documents (average 9.4k words) and summaries (553 words) than other long document datasets, such as ArXiv, PubMed \cite{cohan-etal-2018-discourse}, BillSum \cite{kornilova2019billsum}, and BigPatent \cite{sharma2019bigpatent};

\paragraph{arXiv} \cite{cohan-etal-2018-discourse} is a dataset of scientific articles from arXiv. Abstracts of the articles are used as the target summary. ArXiv is chosen over PubMed \cite{cohan-etal-2018-discourse} as arXiv contains longer articles compared to PubMed. 

A detailed comparison of the datasets used can be found in Table \ref{tab:dataset-comparison}.

% \iffalse
%%%%%%%%%%%%%%%%%%%%%%%%%%%%%%%%%%%%%%%%%%%%%%%%%%%%%%%%%%%%%%%%%%%%%%%%%%%%%%
\begin{table}[t]
\centering
\small
\begin{tabular}{@{}lcccc@{}}
\toprule
Dataset &Query &Format &Src. leng. &Tgt. leng. \\
\midrule
GovReport &\xmark &Doc. &9,409 &553 \\
arXiv &\xmark &Doc. &6,030 &273 \\
QMSum &\cmark &Dial. &9,070 &69 \\
\bottomrule
\end{tabular}
% \fi
% %%%%%%%%%%%%%%%%%%%%%%%%%%%%%%%%%%%%%%%%%%%%%%%%%%%%%%%%%%%%%%%%%%%%%%%%%%%%%%
% \begin{table*}[t]
% \centering
% \small
% \begin{tabular}{@{}lccc@{}}
% \toprule
% Dataset &Src. length &Tgt. length \\
% \midrule
% Document \\
% \quad GovReport \cite{HuangCPJW21} &9409 &553 \\
% \quad PubMed \cite{cohan-etal-2018-discourse} &3049 &202 \\
% \quad ArXiv \cite{cohan-etal-2018-discourse} &6030 &273 \\
% \quad BillSum \cite{kornilova2019billsum} &1813 &208 \\
% \quad BigPatent \cite{sharma2019bigpatent} &3573 &117 \\
% \midrule
% Dialogue \\
% \quad QMSum \cite{ZhongYYZMJACLQR21} &9070 &69 \\
% \quad AMI \cite{carletta2005ami} &6008 &297 \\
% \quad ICSI \cite{janin2003icsi} &13317 &489 \\
% \quad MediaSum \cite{zhu2021mediasum} &1554 &14 \\
% \quad SummScreen \cite{chen2021summscreen} &6613 &337 \\

% \bottomrule
% \end{tabular}
% \caption{\label{tab:dataset-comparison} Comparison of Document and Dialogue Summarization Dataset.}%\ys{I think for QMSum, src is 9070 and target is 69}} 
% \end{table*}
% %%%%%%%%%%%%%%%%%%%%%%%%%%%%%%%%%%%%%%%%%%%%%%%%%%%%%%%%%%%%%%%%%%%%%%%%%%%%%%
\caption{\label{tab:dataset-comparison} Comparison of evaluation benchmarks.} 
\end{table}
%%%%%%%%%%%%%%%%%%%%%%%%%%%%%%%%%%%%%%%%%%%%%%%%%%%%%%%%%%%%%%%%%%%%%%%%%%%%%%

\subsection{Baselines and Implementation}

\paragraph{Baselines for Comparisons} 
We compare \method with the previous \revise{R3}{state-of-the-art} methods on the aforementioned three datasets. More specifically: 1) For GovReport, we report the performance from the original paper, which uses various encoder self-attention and the proposed HEPOS encoder-decoder attention; 2) For QMSum, we compare with \citet{zhong2021dialoglm}, the current SoTA and other baselines mentioned in that work; 3) For arXiv, we include the results from the best performing models in previous works, including ExtSum-LG \cite{xiao2019extractive}, PEGASUS \cite{zhang2020pegasus}, DANCER \cite{Gidiotis2020Dancer}, BigBird \cite{zaheer2020big}, HEPOS + LSH \cite{HuangCPJW21}, HAT-BART \cite{rohde2021hierarchical}, Longformer \cite{Beltagy20Longformer}, and SSN-DM \cite{cui2021sliding}. Note that those baselines spans over different strategies to handle long input, such as sparse-attention (HEPOS, BigBird, Longformer), hierarchical attention (HAT-BART), extract-then-generate (Locator + different generators).

%%%%%%%%%%%%%%%%%%%%%%%%%%%%%%%%%%%%%%%%%%%%%%%%%%%%%%%%%%%%%%%%%%%%%%%%%%%%%%
\begin{table}[t]
\centering
\small
\begin{tabular}{@{}lccc@{}}
\toprule
~  			      & R-1 & R-2 & R-L \\
\midrule
BART(1024) &52.83 &20.50 &50.14 \\
\revise{R3}{\textit{BART w/ sparse attn.}}\\
\quad Stride (4096) &54.29 &20.80 &51.35 \\
\quad LIN. (3072) &44.84 &13.87 &41.94 \\
\quad LSH (4096) &54.75 &21.36 &51.27 \\
\quad Sinkhorn (5120) &55.45 &21.45 &52.48\\
\textit{BART w/ sparse attn. + HEPOS} \\
\quad LSH (7168) &55.00 &21.13 &51.67 \\
\quad Sinkhorn (10240) &56.86 &22.62 &53.82 \\
\midrule
\textsc{Dyle} (dynamic) &\textbf{61.01} &\textbf{28.83} &\textbf{57.82} \\
\bottomrule
\end{tabular}
\caption{\label{tab:automatic-evaluation-govreport} Results on GovReport, where R stands for the ROUGE metric and the number in the brackets denotes maximum input sequence length of the model. 
} 
\end{table}
%%%%%%%%%%%%%%%%%%%%%%%%%%%%%%%%%%%%%%%%%%%%%%%%%%%%%%%%%%%%%%%%%%%%%%%%%%%%%%

%%%%%%%%%%%%%%%%%%%%%%%%%%%%%%%%%%%%%%%%%%%%%%%%%%%%%%%%%%%%%%%%%%%%%%%%%%%%%%
\begin{table}[t]
\centering
\footnotesize
\begin{tabular}{@{}l@{}ccc@{}}
\toprule
~  			      & R-1 & R-2 & R-L \\
\midrule
\revise{R3}{\textit{Locator as extractor}} \\
\quad PGNet (2048) &28.74 &5.98 &25.13 \\
\quad Bart-large (3072) &32.16 &8.01 &27.72 \\
\quad HMNet (8192) &32.29 &8.67 &28.17 \\
\quad Longformer (8192) &31.60 &7.80 &20.50 \\
\quad UNILM-base (5120) &29.14 &6.25 &25.46 \\
\quad UNILM-CP (5120) &29.19 &6.73 &25.52 \\
\textit{UniLM with DialogLM pretraining} \\
\quad DialogLM (5120) &34.02 &9.19 &29.77 \\
\quad DialogLM - Sparse (8192) &33.69 &9.32 &30.01 \\
\midrule
\textsc{Dyle} (dynamic) &\textbf{34.42} &\textbf{9.71} &\textbf{30.10}  \\
\bottomrule
\end{tabular}
\caption{\label{tab:automatic-evaluation-qmsum} Results on QMSum. The baseline performance numbers are from \citet{zhong2021dialoglm}.} 
\end{table}
%%%%%%%%%%%%%%%%%%%%%%%%%%%%%%%%%%%%%%%%%%%%%%%%%%%%%%%%%%%%%%%%%%%%%%%%%%%%%%

%%%%%%%%%%%%%%%%%%%%%%%%%%%%%%%%%%%%%%%%%%%%%%%%%%%%%%%%%%%%%%%%%%%%%%%%%%%%%%
\begin{table}[t]
\centering
\small
\begin{tabular}{@{}l@{}ccc@{}}
\toprule
~  			      & R-1 & R-2 & R-L \\
\midrule
Prior Work \\
\quad ExtSum-LG (dynamic) &44.01 &17.79 &39.09 \\
\quad PEGASUS (3072) &44.21 &16.95 &38.83 \\
\quad DANCER-PEGASUS (dynamic) \ \ &45.01 &17.60 &40.56 \\
\quad BigBird-PEGASUS (3072) &46.63 &19.02 &41.77 \\
\quad LSH (7168) &\textbf{48.24} &\textbf{20.26} &41.78 \\
\quad HAT-BART (3072) &46.68 &19.07 &\textbf{42.17} \\
\quad LED-large (16384) &46.63 &19.62 &41.83 \\
\quad SSN-DM (dynamic) &45.03 &19.03 &32.58 \\
\midrule
\textsc{Dyle} (dynamic) &46.41 &17.95 &41.54 \\
\bottomrule
\end{tabular}
\caption{
\label{tab:automatic-evaluation-arxiv} Results on arXiv.} 
\end{table}

%%%%%%%%%%%%%%%%%%%%%%%%%%%%%%%%%%%%%%%%%%%%%%%%%%%%%%%%%%%%%%%%%%%%%%%%%%%%%%

%%%%%%%%%%%%%%%%%%%%%%%%%%%%%%%%%%%%%%%%%%%%%%%%%%%%%%%%%%%%%%%%%%%%%%%%%%%%%%
\begin{table}[t]
\centering
\small
\begin{tabular}{@{}lccc@{}}
\toprule
~  			      & R-1 & R-2 & R-L \\
\midrule
GovReport \\
\quad Full &\textbf{61.01} &\textbf{28.83} &\textbf{57.82} \\ 
\quad w/o hybrid &60.89 &28.28 &57.31\\
\quad w/o consistency &60.59 &28.48 &57.49\\
\quad w/o oracle &57.57 &25.92 &53.14\\
\midrule
QMSum                  \\
\quad Full &\textbf{34.42} &\textbf{9.71} &\textbf{30.10}  \\
\quad w/o hybrid &31.77 &8.33 &28.37 \\
\quad w/o consistency &32.51 &8.77 &28.94 \\
\quad w/o oracle &32.13 &8.38 &28.63\\
\bottomrule
\end{tabular}
\caption{\label{tab:ablation-study} Ablation study for auxiliary optimizations.}
\end{table}
%%%%%%%%%%%%%%%%%%%%%%%%%%%%%%%%%%%%%%%%%%%%%%%%%%%%%%%%%%%%%%%%%%%%%%%%%%%%%%

%%%%%%%%%%%%%%%%%%%%%%%%%%%%%%%%%%%%%%%%%%%%%%%%%%%%%%%%%%%%%%%%%%%%%%%%%%%%%%
\begin{table*}[t]
\centering
\small
\begin{tabular}{@{}llccccccccc@{}}
\toprule
~ & ~ & \multicolumn{3}{c}{ROUGE-1} & \multicolumn{3}{c}{ROUGE-2} & \multicolumn{3}{c}{ROUGE-L} \\
\cmidrule(r){3-5} \cmidrule(lr){6-8} \cmidrule(l){9-11}
~ & ~ & P & R & F1 & P & R & F1 & P & R & F1 \\
\midrule
\multirow{2}*{GovReport} & Extracted snippets &48.98 &\textbf{73.40} &57.56 &24.20 &\textbf{36.59} &28.53 &46.28 &\textbf{69.25} & 54.35 \\
& Generated summaries &\textbf{63.16} &61.61 & \textbf{61.01} &\textbf{29.85} &29.10 &\textbf{28.83} &\textbf{59.88} &58.35 & \textbf{57.82} \\
\midrule
\multirow{2}*{QMSum} & Extracted snippets & 4.25 & \textbf{76.90} & 7.74 & 1.36 & \textbf{28.41} & 2.49 & 3.99 & \textbf{72.83} & 7.26 \\
& Generated summaries & \textbf{29.78} & 45.64 &\textbf{34.42} & \textbf{8.39} & 13.06 & \textbf{9.71} & \textbf{26.14} & 39.70 &\textbf{30.10}  \\
\bottomrule
\end{tabular}
\caption{\label{tab:extracted-snippets-interpretability} Precision-recall decomposition of ROUGE scores of extracted snippets and generated summaries.}
\end{table*}
%%%%%%%%%%%%%%%%%%%%%%%%%%%%%%%%%%%%%%%%%%%%%%%%%%%%%%%%%%%%%%%%%%%%%%%%%%%%%%

\subsection{Implementation Details}
\paragraph{Pretrained-LM} The extractor is initialized with \texttt{RoBERTa-base} \cite{DBLP:journals/corr/abs-1907-11692} weights. The generator is initialized with \texttt{BART-large} \cite{LewisLGGMLSZ20} weights. We use the Adam optimizer and set the extractor learning rate to 5e-5 and the generator learning rate to 5e-6. 
\paragraph{Hyperparameters} $\lambda_{\itm{g}}$, $\lambda_{\itm{o}}$, and $\lambda_{\itm{c}}$ are the coefficients for the generation loss, oracle loss, and the consistency loss respectively. 
For $\lambda_{\itm{g}}$ and $\lambda_{\itm{o}}$, we did a 2-step binary search between 0 and 2. For $\lambda_{\itm{c}}$, we did a 3-step binary search between 0 and 10. For the QMSum dataset, we used $\lambda_{\itm{g}} = 1$, $\lambda_{\itm{o}} = 1$,  $\lambda_{\itm{c}} = 1$. For the GovReport dataset, we used $\lambda_{\itm{g}} = 0.5$, $\lambda_{\itm{o}} = 1$,  $\lambda_{\itm{c}} = 1$. For the ArXiv dataset, we used $\lambda_{\itm{g}} = 0.5$, $\lambda_{\itm{o}} = 1$,  $\lambda_{\itm{c}} = 5$.
\paragraph{Hardware} We apply gradient checkpointing \cite{DBLP:journals/corr/ChenXZG16} to save the GPU memory. Each experiment is run on one NVIDIA Quadro RTX 8000 GPU. The effective batch size is set to 8.

\section{Experiment Results}

\subsection{Main Results}
\label{subsec:main-results}
The evaluation results are summarized in Table \ref{tab:automatic-evaluation-govreport}, Table \ref{tab:automatic-evaluation-qmsum}, and Table \ref{tab:automatic-evaluation-arxiv}. 
For GovReport, \textsc{Dyle} yields gains of 4.15/6.21/4.00 of ROUGE-1/2/L scores compared to the previous best method. \revise{R3}{Experiments on GovReport show that \textsc{Dyle} is performant over prior sparse attention approaches.}

On QMSum, \textsc{Dyle} yields the new state-of-the-art ROUGE-1/2/L scores of 34.42/9.71/30.10, outperforms UniLM with DialogLM pretraining. 
\revise{R3}{Comparing \textsc{Dyle} with locator-based models on the QMSum dataset shows that \textsc{Dyle} outperforms prior extract-then-generate approaches where the locator is independently trained with intermediate annotated text spans.} This shows the effectiveness of \textsc{DYLE}'s joint training approach. These results show that \textsc{Dyle} can be applied to both the long document summarization and long dialogue summarization tasks. \method's better performance can be attributed to lowered information loss between the extraction and the generation steps and its ability to handle input of a much longer length.

We notice that while \textsc{Dyle} largely outperforms the LSH baseline \cite{HuangCPJW21} on the GovReport dataset, it underperforms the LSH baseline on arXiv. We posit two reasons. First, the input of the GovReport is much longer than that of arXiv. Most, if not all, of the sentences in the arXiv input article can be processed by the LSH model. 
Second, the summaries of the arXiv dataset are more abstractive than those of GovReport. It is possible that individually extracted text snippet is not the best linguistic unit for generating output tokens. It is our future work to explore the optimal input unit for an extract-then-generate approach. Nevertheless, \textsc{DYLE} outperforms other \revise{}{extraction-based approaches} (\eg SSN-DM \cite{cui2021sliding}) and divide-and-conquer approaches (\eg DANCER \cite{Gidiotis2020Dancer}).

\subsection{Evaluation of Auxiliary Optimizations}
\label{subsec:eval_of_auxiliary_optimization}
We conduct ablation studies to investigate the effectiveness of the auxiliary optimizations we introduced. Specifically, we report the full model's performance after removing 1) hybrid training, 2) consistency loss, 3) extractive oracle loss. 
In our default model, the consistency loss is computed on the combination of the extracted snippets and oracle snippets; in the ``w/o hybrid'' experiment, the consistency loss is only computed on the set of oracle snippets; in ``w/o consistency'' experiment, the consistency loss is not computed. 
The results are summarized in Table \ref{tab:ablation-study}. Note that without the hybrid training optimization, only the extractive oracles will be used to train the generator. 
When the consistency loss is not calculated, the extractor and the generator can be viewed as being trained independently with the extractive oracles.

We see that excluding either of the hybrid training, consistency loss, or oracle loss optimization leads to a performance drop. Training the model without the supervision of the oracle leads to the greatest decrease in model performance, showing the importance of good supervision for the extractor. Removing the consistency loss also decreases the model performance. This shows that the consistency loss allows the extractor to better learn to select salient snippets from the input text and enables \method to generalize better to the test set.

%%%%%%%%%%%%%%%%%%%%%%%%%%%%%%%%%%%%%%%%%%%%%%%%%%%%%%%%%%%%%%%%%%%%%%%%%%%%%%
\begin{figure*}[t]
	\centering
	\subfigure[QMSum sample 1]{
        \includegraphics[width=0.22\linewidth]{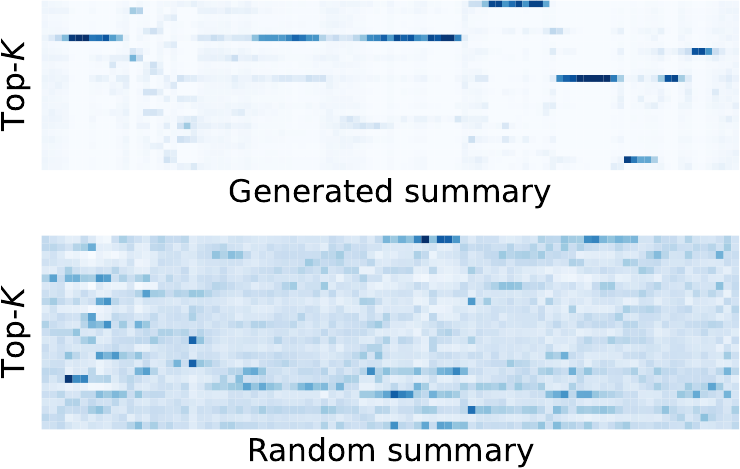}
    }
    \subfigure[QMSum sample 2]{
        \includegraphics[width=0.22\linewidth]{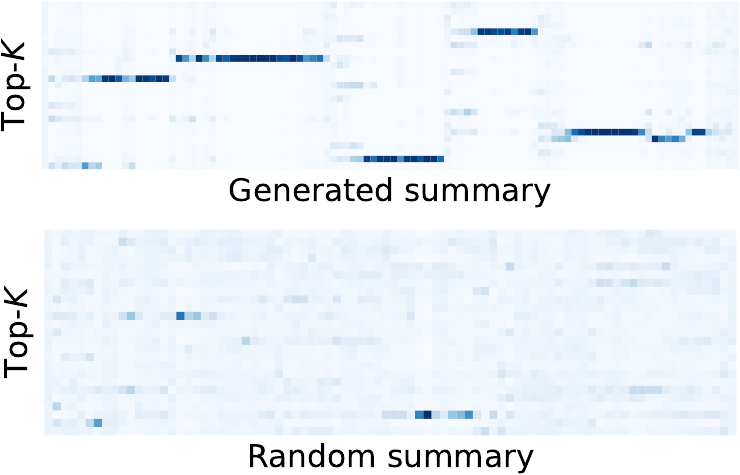}
    }
    \subfigure[QMSum sample 3]{
        \includegraphics[width=0.22\linewidth]{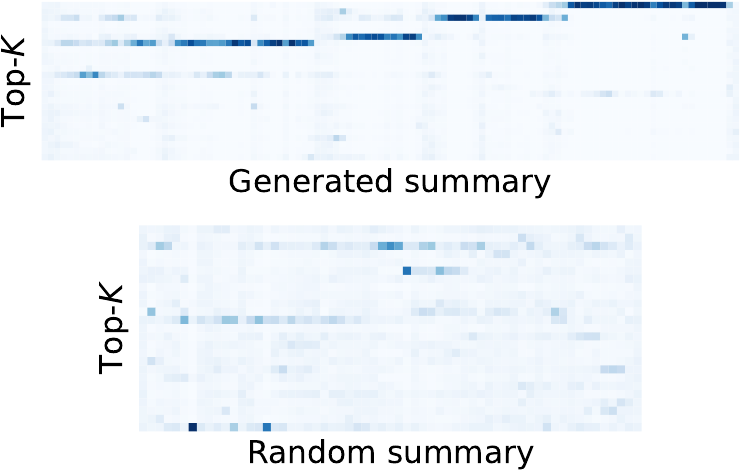}
    }
    %\hspace{0.05\linewidth}
    \subfigure[QMSum sample 4]{
        \includegraphics[width=0.22\linewidth]{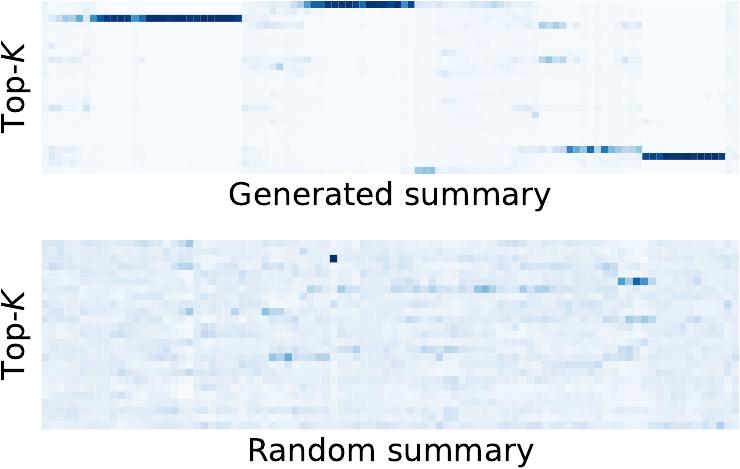}
    }
    \subfigure[GovReport sample 1]{
        \includegraphics[width=0.97\linewidth]{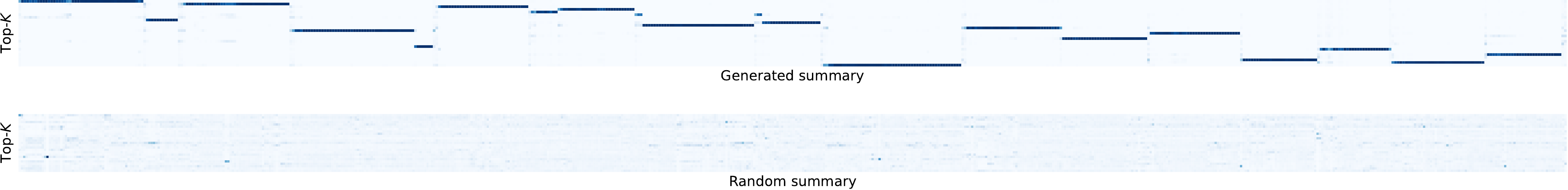}
    }
    \subfigure[GovReport sample 2]{
        \includegraphics[width=0.97\linewidth]{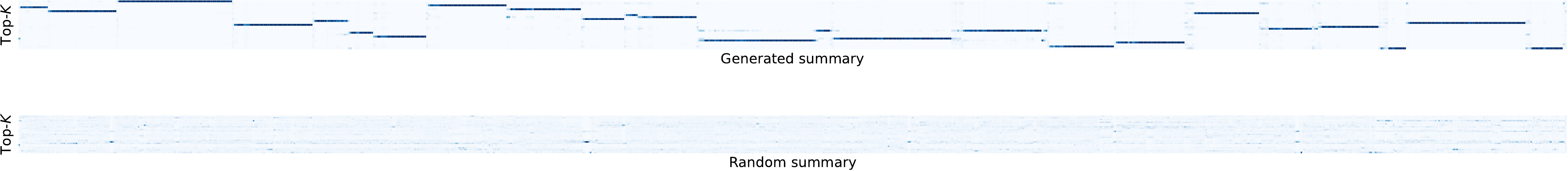}
    }
    \subfigure[GovReport sample 3]{
        \includegraphics[width=0.95\linewidth]{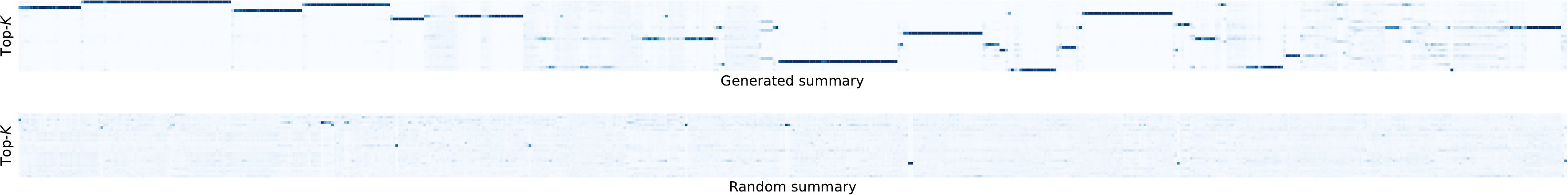}
    }
    \subfigure[GovReport sample 4]{
        \includegraphics[width=0.95\linewidth]{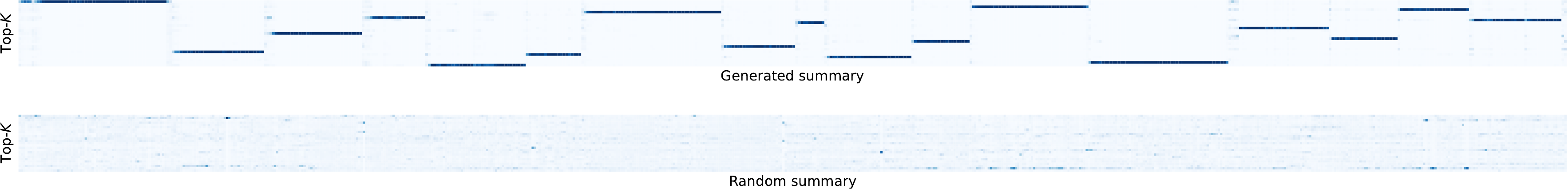}
    }
	\caption{\label{fig:visualization-of-weights} Dynamic weight visualization. We visualized the dynamic weight matrices of the \textit{generated summary} and a \textit{random summary} from other samples in the validation set. $x$-axis: decoding time step; $y$-axis: index of the extracted top-$K$ snippets. Darker squares stand for higher weights. More examples can be found in Appendix \ref{app:dynamic-weights}.
	}
\end{figure*}
%%%%%%%%%%%%%%%%%%%%%%%%%%%%%%%%%%%%%%%%%%%%%%%%%%%%%%%%%%%%%%%%%%%%%%%%%%%%%%

\section{Analysis and Discussion}

\paragraph{Analysis of extracted snippets}
\label{subsec:analaysis_of_extracted_snippets}
We are interested in the amount of salient information passed to the generator. To investigate this, we report the decomposed precision and recall of ROUGE scores in Table~\ref{tab:extracted-snippets-interpretability}. We observe that the extracted snippets have much higher recall than the generated summaries, while the generated summaries have higher precision. This suggests that to improve the overall performance, we can increase the information coverage (i.e., recall) of the extractor and improve the accuracy of the generator in identifying the salient snippets (i.e., precision).

\paragraph{Interpretability of dynamic weights}
Our approach is more interpretable than sparse attention and two-step extraction-generation pipeline methods. Specifically, \textit{dynamic weights} in the generator shows how the information is used throughout the decoding process. In Figure~\ref{fig:visualization-of-weights}, we visualize the dynamic weights for the extracted snippets assigned by the generator during decoding. In each subfigure, we visualize the dynamic weight matrices of the \textit{generated summary} and a \textit{random summary} from other samples in the validation set. The $x$-axis and $y$-axis represent the decoding time step and the index of the extracted top-$K$ snippets, respectively. Darker squares denote higher weights. For each generated summary, we observe multiple consecutive high-weight areas, indicating alignments between the extracted snippets and the generated summary. In contrast, weights are uniformly distributed for random summaries.
Interestingly, we observe that, on QMSum, fewer sentences are considered when generating the summaries. Our explanation for this observation is that QMSum is a query-based dataset, where the queried information is more concentrated in a few snippets. By contrast, we find that a larger number of snippets are used on the GovReport dataset as seen in Figure~\ref{fig:visualization-of-weights}, as GovReport is a general summarization dataset.

\paragraph{Effect of number of extracted snippets}

To evaluate the effect of number of extracted snippets on model performance, we vary the value of $K$ of top-$K$ in Eq.~(\ref{eq:topk}) and test it on both the GovReport and QMSum datasets. We observe that the model performance generally increases as the value of $K$ increases. This is expected as more extracted snippets provide the generator with more information to form a final summary. The results are summarized in Table~\ref{tab:k-values-comparison}. Due to the limit of GPU memory, the largest $K$ value we tried is 25. 

\paragraph{Effect of consistency loss}

We evaluate the effect of consistency loss on extractor performance. Note that removing the consistency loss means that the extractor and the generator are independently trained. The results are presented in Table~\ref{tab:ablation-study} as part of the ablation study. Removing the consistency loss leads to worse model performance. We observe that the consistency loss helps the model better learn the importance of the selected text snippets useful for the generation. 

\paragraph{Extractor performance compared with extractive oracles}
We feed the extractive oracles to the generator. The results are summarized in Table \ref{tab:oracle-test-comparison}. We observe that extractive oracles contain more salient information than the text snippets extracted by the extractor. Feeding the extractive oracle to the generator indicates the upper bound of the extractor performance. However, we observe that the gap between the performance of using the extractive oracle and using the extractor output is relatively small.

%%%%%%%%%%%%%%%%%%%%%%%%%%%%%%%%%%%%%%%%%%%%%%%%%%%%%%%%%%%%%%%%%%%%%%%%%%%%%%
\begin{table}[t]
\centering
\small
\begin{tabular}{@{}lccc@{}}
\toprule
~  			      & R-1 & R-2 & R-L \\
\midrule
GovReport                  \\
\quad $K$=25 &\textbf{61.01} &\textbf{28.83} &\textbf{57.82}  \\
\quad $K$=20 &59.25 &27.46 &55.74\\
\quad $K$=15 &58.55 &26.95 &54.89\\
\quad $K$=10 &54.98 &24.10 &51.25\\
\midrule
QMSum                  \\
\quad $K$=25 &\textbf{34.42} &\textbf{9.71} &\textbf{30.10}  \\
\quad $K$=20 &33.10 &8.69 &29.62\\
\quad $K$=15 &31.78 &8.36 &28.31\\
\quad $K$=10 &33.30 &9.18 &29.53\\
\bottomrule
\end{tabular}
\caption{\label{tab:k-values-comparison} Comparing model performance with different values of $K$ on the GovReport and QMSum dataset} 
\end{table}
%%%%%%%%%%%%%%%%%%%%%%%%%%%%%%%%%%%%%%%%%%%%%%%%%%%%%%%%%%%%%%%%%%%%%%%%%%%%%%

%%%%%%%%%%%%%%%%%%%%%%%%%%%%%%%%%%%%%%%%%%%%%%%%%%%%%%%%%%%%%%%%%%%%%%%%%%%%%%
\begin{table}[t]
\centering
\small
\begin{tabular}{@{}lccc@{}}
\toprule
~  			      & R-1 & R-2 & R-L \\
\midrule
GovReport                  \\
\quad Extractor Output  &61.01 &28.83 &57.82  \\
\quad Oracle &\textbf{68.02} &\textbf{39.16} &\textbf{65.29} \\
\midrule
QMSum                  \\
\quad Extractor Output &34.42 &9.71 &30.10  \\
\quad Oracle &\textbf{39.80} &\textbf{14.74} &\textbf{36.06} \\
\bottomrule
\end{tabular}
\caption{\label{tab:oracle-test-comparison} Feeding extractive oracles to generator. \revise{R2}{"Oracle" is computed based on the gold summary; thus, it is a soft upper-bound of the extractor's performance.}} 
\end{table}
%%%%%%%%%%%%%%%%%%%%%%%%%%%%%%%%%%%%%%%%%%%%%%%%%%%%%%%%%%%%%%%%%%%%%%%%%%%%%%

\paragraph{Comparison with RAG}
The generator of our method is related to but differs significantly from Retrieval-Augmented Generation (RAG) \cite{LewisPPPKGKLYR020}. The similarity only lies in the idea of marginalization over a set of text snippets, which is shown to be useful in question answering as well \cite{ni-etal-2021-mitigating}. However, unlike our \textit{dynamic} weights, the weights in RAG remains \textit{static} during decoding. In our notations, RAG's generation probability can be formulated as:
\begin{equation}
\label{eq:rag}
\begin{split}
    &P_{\theta}(y|q, X_{K}) = \prod_{t=1}^{T} P_{\theta}(y_{t}|q, X_{K}, y_{<t}) \\ 
    &\ \ \ = \prod_{t=1}^{T} \sum_{x \in X_{K}}P_{\theta}(y_{t}|q, x, y_{<t}) P_{\theta}(x|q, X_{K})
\end{split}
\end{equation}
The static weight $P_{\theta}(x|q, X_{K})$ in Eq.~\ref{eq:rag} is computed based on $q$ and $X_{K}$, while our dynamic weight $P_{\theta}(x|q, X_{K}, y_{<t})$ is additionally conditioned on the already generated tokens. 

\paragraph{Limitations and future directions} 
We acknowledge that joint training of the extractor and the generator cannot eliminate information loss, which might be addressed by combining \textsc{Dyle} and sparse attention to encode longer snippets. Though formulated for long-input summarization, \textsc{Dyle} can be applied to general long-input generation tasks where information is scattered across the input, e.g., open-domain question answering and multi-turn dialogue systems with long dialogue history. 

\section{Conclusions}
In this paper, we propose the first framework that jointly trains an extract-then-generate model with latent extraction. The first-step extraction picks out salient information from the long input, thereby extending the input length that the model can handle. Our novel joint training method addresses the challenge of information loss associated with the prior extract-then-generate approaches. Our model largely outperforms the current state-of-the-art on GovReport and QMSum, while achieving strong results on arXiv. Lastly, \textsc{Dyle} has the advantages of being able to process arbitrarily long input with a lower memory cost and interpretable generator weights. 

% Commented for submission. 
\section*{Acknowledgment}
The authors would like to thank Yixin Liu and Ming Zhong for the discussions. We also would like to thank the anonymous reviewers for their helpful comments. This work is supported in part by a grant from Microsoft Research. 

\clearpage
\bibliography{anthology,custom}
\bibliographystyle{acl_natbib}

\appendix

\section{Additional Dynamic Weight Visualization}
\label{app:dynamic-weights}

%%%%%%%%%%%%%%%%%%%%%%%%%%%%%%%%%%%%%%%%%%%%%%%%%%%%%%%%%%%%%%%%%%%%%%%%%%%%%%
\begin{figure*}[h]
	\centering
	\subfigure[QMSum sample 1]{
        \includegraphics[width=0.25\linewidth]{qmsum/vis0.pdf}
    }
    \subfigure[QMSum sample 2]{
        \includegraphics[width=0.25\linewidth]{qmsum/vis1.pdf}
    }
    \subfigure[QMSum sample 3]{
        \includegraphics[width=0.25\linewidth]{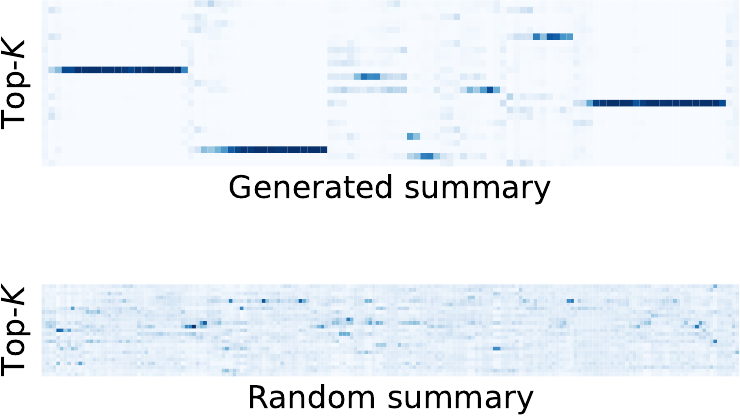}
    }
    \subfigure[QMSum sample 4]{
        \includegraphics[width=0.25\linewidth]{qmsum/vis3.pdf}
    }
    \subfigure[QMSum sample 5]{
        \includegraphics[width=0.25\linewidth]{qmsum/vis4.pdf}
    }
	\caption{\label{fig:more-visualization-of-weights-qmsum} Dynamic weights visualization on QMSum. }
\end{figure*}
%%%%%%%%%%%%%%%%%%%%%%%%%%%%%%%%%%%%%%%%%%%%%%%%%%%%%%%%%%%%%%%%%%%%%%%%%%%%%%

%%%%%%%%%%%%%%%%%%%%%%%%%%%%%%%%%%%%%%%%%%%%%%%%%%%%%%%%%%%%%%%%%%%%%%%%%%%%%%
\begin{figure*}[h]
	\centering
	\subfigure[GovReport sample 1]{
        \includegraphics[width=0.95\linewidth]{govreport/vis0.pdf}
    }
    \subfigure[GovReport sample 2]{
        \includegraphics[width=0.95\linewidth]{govreport/vis1.pdf}
    }
    \subfigure[GovReport sample 3]{
        \includegraphics[width=0.95\linewidth]{govreport/vis2.pdf}
    }
    \subfigure[GovReport sample 4]{
        \includegraphics[width=0.95\linewidth]{govreport/vis3.pdf}
    }
    \subfigure[GovReport sample 5]{
        \includegraphics[width=0.95\linewidth]{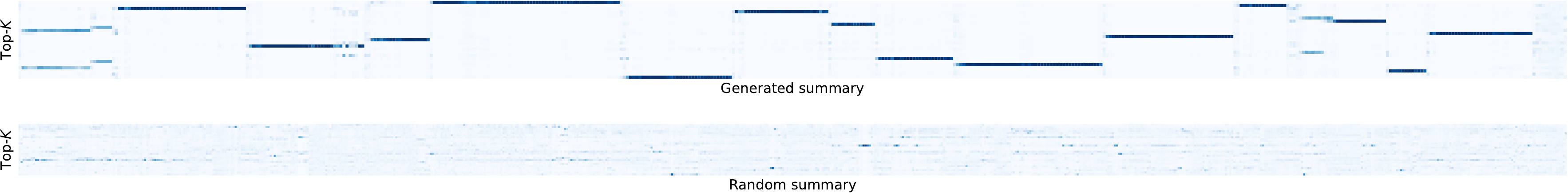}
    }
	\caption{\label{fig:more-visualization-of-weights-govreport} Dynamic weights visualization on GovReport. }
\end{figure*}
%%%%%%%%%%%%%%%%%%%%%%%%%%%%%%%%%%%%%%%%%%%%%%%%%%%%%%%%%%%%%%%%%%%%%%%%%%%%%%

\end{document}